\documentclass{article}

\usepackage{microtype}
\usepackage{graphicx}
\usepackage{subfigure}
\usepackage{booktabs} %
\usepackage[table,xcdraw]{xcolor}
\usepackage{paralist}

\usepackage{hyperref}

\usepackage[accepted]{icml2020}

\usepackage{amsmath,amsfonts,bm,amsthm}

\def\eqref#1{equation~\ref{#1}}

\def\1{\bm{1}}

\DeclareMathAlphabet{\mathsfit}{\encodingdefault}{\sfdefault}{m}{sl}
\SetMathAlphabet{\mathsfit}{bold}{\encodingdefault}{\sfdefault}{bx}{n}

\DeclareMathOperator*{\argmax}{arg\,max}

\newtheorem{theorem}{Theorem}
\newtheorem{hypothesis}[theorem]{Hypothesis}
\newtheorem*{observation}{Observation}

\icmltitlerunning{An Interpretability Illusion for BERT}

\begin{document}

\twocolumn[
\icmltitle{An Interpretability Illusion for BERT}

\icmlsetsymbol{equal}{*}

\begin{icmlauthorlist}
\icmlauthor{Tolga Bolukbasi}{equal,goo}
\icmlauthor{Adam Pearce}{equal,goo}
\icmlauthor{Ann Yuan}{equal,goo}
\icmlauthor{Andy Coenen}{goo}
\icmlauthor{Emily Reif}{goo}
\icmlauthor{Fernanda Viégas}{goo}
\icmlauthor{Martin Wattenberg}{goo}
\end{icmlauthorlist}

\icmlaffiliation{goo}{Google Research, Cambridge, MA, USA}

\icmlcorrespondingauthor{Tolga Bolukbasi}{tolgab@google.com}
\icmlcorrespondingauthor{Adam Pearce}{adampearce@google.com}
\icmlcorrespondingauthor{Ann Yuan}{annyuan@google.com}

\icmlkeywords{Machine Learning, BERT, NLP, Interpretability}

\vskip 0.3in
]

\printAffiliationsAndNotice{\icmlEqualContribution} %

\begin{abstract}
We describe an ``interpretability illusion'' that arises when analyzing the BERT model. Activations of individual neurons in the network may spuriously appear to encode a single, simple concept, when in fact they are encoding something far more complex. The same effect holds for linear combinations of activations. We trace the source of this illusion to geometric properties of BERT's embedding space as well as the fact that common text corpora represent only narrow slices of possible English sentences. We provide a taxonomy of model-learned concepts and discuss methodological implications for interpretability research, especially the importance of testing hypotheses on multiple data sets. 

\end{abstract}

\section{Introduction}
\label{sec:introduction}

An outstanding problem in the field of neural networks is understanding how they represent meaning. One simple hypothesis is that the activation level of an individual unit in a network encodes the presence or absence of a meaningful concept. A generalization of this idea suggests that concepts are encoded by linear combinations of neural activations.

This point of view has proved fruitful in analyzing image networks \cite{bau2017network, olah2020zoom, kim2018interpretability}. 
Could the same type of analysis uncover the key concepts that are learned by neural language networks?

This paper describes a surprising phenomenon, a kind of ``interpretability illusion,'' that arose from such an experiment on BERT, a model for general language representations \cite{devlin2018bert}. The basic set-up for our experiment (described in detail in Section \ref{sec:methodology}) was designed to probe whether individual neurons in BERT might have human-interpretable meaning. We gave BERT a large dataset of sentences as inputs, chose a target neuron from the final layer, and examined the sentences that maximally activated it. Our intuition was that if these maximally-activating sentences shared a common pattern, it would be an indication that the target neuron had learned to detect this pattern.

Indeed, many of the neurons we probed did show strong, consistent patterns of activation. For example, of the 164,246 sentences in the Quora Question Pairs dataset \cite{WinNT}, here are three of the sentences which activate neuron 221 in layer 12 most strongly:
\begin{compactitem}
    \item \textit{"What is the meaning behind the song ""Angel"" by Eric Clapton?"}
    \item \textit{"What's the meaning of Johnny Cash's song ""King of the Hill""?"}
    \item \textit{"What is the meaning behind the Tears for Fears song ""Mad  World"",  such  as  the  lyric, ""All  around  me  are familiar faces""?"}
\end{compactitem}
These strongly suggest that neuron 221 encodes a concept related to song titles, or perhaps the very specific syntactic structure of these sentences. Based on this evidence alone, it might be tempting to conclude that one can easily interpret the meanings of individual neurons in the final layer of BERT.

The plot thickened, however, when we tried the same set of experiments with the same model (no fine-tuning or other modifications) but a different dataset. Using as input a question answering dataset drawn from Wikipedia (58,645 sentences), the top activating sentences for neuron 221 had nothing to do with the meanings of song titles. Instead, they included:

\begin{compactitem}
    \item \textit{On 16 June 2006, it was announced that Everton had entered into talks with Knowsley Council and Tesco over the possibility of building a new 55,000 seat stadium, ex-pandable to over 60,000, in Kirkby.}
    \item \textit{On 15 September 1940, known as the Battle of Britain Day, an RAF pilot, Ray Holmes of No. 504 Squadron RAF rammed a German bomber he believed was going to bomb the Palace.}
    \item \textit{On 20 August 2010, Queen's manager Jim Beach put out a Newsletter stating that the band had signed a new contract with Universal Music.}
\end{compactitem}

Here we are forced to reconsider our initial interpretation. Based on this dataset, it would seem that neuron 221 encodes historical events, or perhaps sentences beginning with a date. 

In search of tie-breaking evidence, we repeated our analysis on a sample (198,085 sentences) from the Toronto BookCorpus dataset. The top activating sentences there included:

\begin{compactitem}
    \item \textit{Lara pulled out the document Reed had supplied from Gresham's briefcase.}
    \item \textit{I take Kellan's business card from my pocket and stretch it over to Realm.}
    \item \textit{Pilcher took a walkie-talkie out of his coat and spoke into the receiver.}
\end{compactitem}

The picture is now more complicated. The BookCorpus dataset suggests yet a third interpretation for neuron 221. Note that there's nothing special about neuron 221: many other neurons show similar behavior. What we have seen is that for each data set, looking at maximally-activating sentences produces consistent, interpretable patterns for many neurons. These patterns, however, are \textit{not} consistent across datasets. We consider this to be an \textit{interpretability illusion}. The fact that a seemingly consistent pattern can turn out to be a mirage has clear implications for interpretability research. 

In this paper, we provide evidence that this illusion is a general, reproducible phenomenon of the BERT model. We also consider several possible explanations for the illusion, suggesting that it is useful to separate the notion of \textit{dataset-level} concept directions from global concept directions. 

To summarize our main contributions:
\begin{compactenum}
    \item We identify an interpretability illusion that arises when analyzing a language model's activation space.
    \item We provide the recommendation that interpretability researchers conduct their experiments on multiple datasets.
    \item We investigate possible causes of the illusion, using a taxonomy of the geometric properties of model-learned concepts: local, global, and dataset-level concept directions. %
\end{compactenum}

\section{Related work}
There is a large body of prior work exploring the embedding spaces of language models. These embedding spaces exhibit both local and global structure. There is local structure in that the nearest neighbors of a datapoint's embedding are similar to it \cite{bengio2003neural}. Global directions in embedding spaces have also been found to encode specific concepts \citep{mikolov2013efficient, bolukbasi2016, raghu2017svcca, olah2020zoom, vig2020causal, DBLP:journals/corr/LiCHJ15}. In the case of language models, these directions combine to form representations that enable sophisticated language processing (\citet{manning2020}). Indeed, following the probing method of \citet{tenney2019what}, \citet{dalvi2020b} finds that different elements of linguistic understanding can be localized to individual or small groups of neurons. 

There has been substantial work (both for images and text) on determining what a specific neuron encodes by considering which inputs maximally activate it. There are two main ways of doing this: first, by generating such inputs, which can provide clues regarding the encoded pattern \citep{nguyen2016, poerner-etal-2018-interpretable, bauerle_wexler}, and second, by looking for patterns among real samples that maximally activate a neuron, where the samples are drawn from some preexisting dataset (\citet{na2019}). 

Our paper focuses on the latter approach. To review a few more examples, \citet{zhou2015} and \citet{bau2017network} use this technique to find neurons that respond to particular objects in natural scenes. \citet{dalvi2019} look at language models, comparing the concepts for a neuron found by this technique to those found by probing. \citet{szegedy2014} find that convolutional networks contain neurons that activate in response to semantically related inputs. \citet{zeiler2014visualizing} obtain a similar result by visualizing the top activating images patches for a given feature. \citet{olah2017feature} compare the maximally activating images from a dataset with images generated to maximize that neuron. %

There is also a body of work using concept directions for measuring and mitigating unwanted biases in models \citep{DBLP:journals/corr/abs-1906-00742, DBLP:journals/corr/abs-1904-04047, bolukbasi2016}. Our experiments suggest that using these techniques for sentence models without validating the directions on multiple datasets could have unintended effects.

\section{Establishing the illusion}
\label{sec:establishing_illusion}

Motivated by this body of work, we set out to find meaningful neurons and concept directions in BERT's activation space. Our pilot experiments with a single test dataset suggested clear, consistent meanings for many neurons. However, as described in the introduction, many of those interpretations disappeared when we tried to confirm them with a different data set. 

\subsection{Datasets}
\label{sec:data}

To test the extent of the illusion, we base our experiments on four different text corpora.

\begin{compactenum}
    \item Quora Question Pairs (\textbf{QQP}): QQP contains questions from the question-answering website Quora, with 164,246 datapoints. \cite{WinNT}.
    \item Question-answering Natural Language Inference (\textbf{QNLI}): QNLI contains passages from Wikipedia, with 58,645 datapoints. \cite{wang2019glue}.
    \item Wikipedia (\textbf{Wiki}): Wiki contains a random subset of English Wikipedia as prepared in \cite{devlin2018bert},  203,736 datapoints.
    \item Toronto BookCorpus (\textbf{Books}): Books contains sentences from online novels. We sampled 198,085 sentences from the original data set. \cite{zhu2015aligning}.
\end{compactenum}

\subsection{Experiments}
\label{sec:methodology}

We began by creating embeddings for the 624,712 sentences in our four datasets. To do this, we used the BERT-base uncased model from the HuggingFace Transformers library with no fine tuning or dataset specific modifications. We used the final layer hidden state of each sentence's $[CLS]$ token as its embedding. 

\begin{figure}[!t]
    \centering
    \includegraphics[width=1.0\columnwidth]{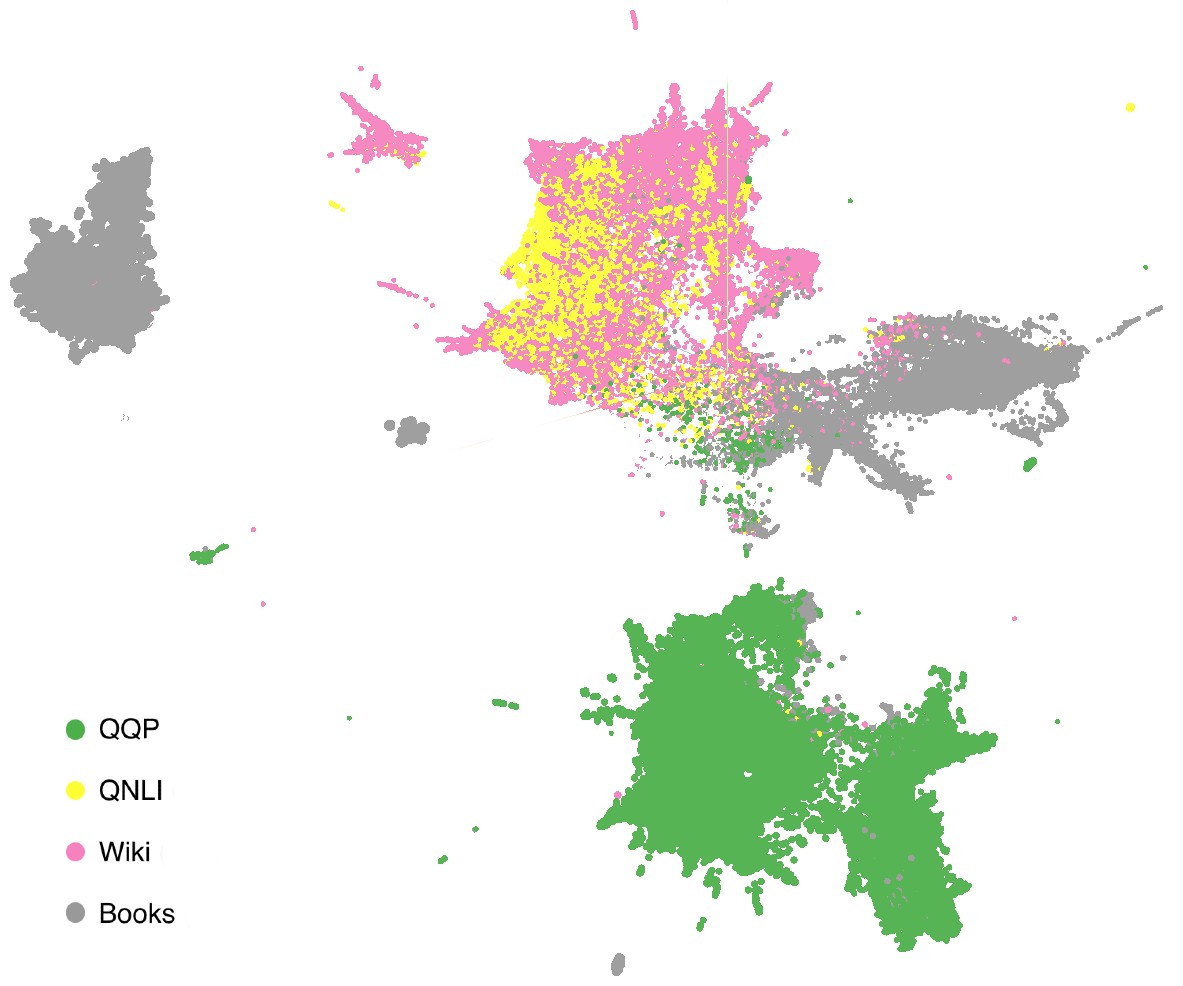}
    \caption{Visualization of sentence embeddings from the QQP, QNLI, Wiki, and Books datasets using UMAP showing that the four datasets form distinct clusters. Our methodology for extracting these embeddings is described in Section \ref{sec:methodology}.}
    \label{fig:umap}
\end{figure}

Several methods for extracting aggregate sequence representations from BERT can be found in the literature \cite{reimers2019}. However our method remains the default for HuggingFace pipelines, and is used in the original BERT paper. We also built an exploratory visualization (Figure \ref{fig:umap}) demonstrating that these embeddings give rise to highly coherent clusters across scales, giving us additional confidence in their representational validity.

We then randomly picked a set of neurons and looked at their top activating sentences from each dataset. For convenience, we identify a neuron with a basis vector in BERT's 768-dimensional embedding space; that is, a one-hot vector $x(d) \in R^k$ where:

\begin{equation}
    x(d)_l=
    \left\{\begin{array}{lr}
      1, & \text{if } l=d \\
      0, & \text{otherwise}
    \end{array}\right.
\end{equation}
For each neuron we find its top activating sentences by sorting sentence embeddings according to their activation level, meaning the dot product with this vector.

We also find top activating sentences for a set of random directions, rather than basis vectors. In this case we sort sentence embeddings according to their inner product with the random direction. Specifically, for a dataset $ S $ and a vector $ v $: 
\begin{equation}
    \label{eqn:top_activating}
    \text{Top activating sentence for } v = \argmax_{x \in \mathcal{S}} \langle x, v \rangle\
\end{equation}

In this paper, we will refer to the dot product between a sentence embedding and a direction as their \textit{projection score}.

Next we built an annotation interface that optionally shows: (1) the top ten activating sentences for a neuron, (2) the top ten activating sentences for a random direction, or (3) a random set of ten sentences. We annotated whether each set of ten sentences contained a pattern, and if so, which sentences demonstrated the pattern. During annotation we knew which dataset the sentences were drawn from, but \textit{not} whether the sentences were top activating (conditions (1), (2)) or randomly drawn (condition (3)). In total we annotated 25 neurons (randomly selected), 33 random directions, and 29 random sets of sentences (Table \ref{tab:meaning_counts}).

To define our notion of pattern: a pattern is simply a property shared by a set of sentences. The property may be structural, i.e. the sentences are all the same length. It may also be lexical, i.e. the sentences all contain some variant of the phrase 'coat of arms'. We use these patterns as proxies for learned concepts by the model.

\begin{table*}[tb]
\centering
\resizebox{\linewidth}{!}{%
\setlength{\tabcolsep}{1pt}
\renewcommand{\arraystretch}{1.3}
\tiny
\begin{tabular}{l|ll|ll|ll|ll|lll}
                          & \textbf{QQP}        &                      & \textbf{QNLI}       &                      & \textbf{Wiki}       &                      & \textbf{Books}      &      & \textbf{All} &  &                     \\                  
                          \textbf{Contains patterns?} & \textbf{Yes} & \textbf{No} & \textbf{Yes} & \textbf{No} & \textbf{Yes} & \textbf{No} & \textbf{Yes} & \textbf{No} & \textbf{Yes} & \textbf{No} & \textbf{Conflicting} \\ \midrule
\textbf{Neurons}     & 3 (60\%)            & 1 (20\%)             & 8 (100\%)           & 0                    & 6 (100\%)           & 0                    & 3 (50\%)            & 2 (33\%)   &  20 (80\%)           & 3 (12\%)   & 2 (8\%)            \\
\textbf{Random direction} & 10 (100\%)          & 0                    & 10 (83\%)           & 0                    & 5 (100\%)           & 0                    & 2 (33\%)            & 0        & 27 (82\%)           & 0          & 6 (18\%)        \\
\textbf{Random sentences}           & 0                   & 5 (100\%)            & 2 (22\%)            & 3 (33\%)             & 1 (11\%)            & 3 (33\%)             & 1 (17\%)            & 3 (50\%)      & 4 (14\%)            & 14 (48\%)           & 11 (38\%)      
\end{tabular}} 
\caption{\label{tab:meaning_counts} Annotation results for each dataset. For individual datasets, the remaining percentage is for conflicting annotations.}
\bigskip
\resizebox{\linewidth}{!}{%
\begin{tabular}{p{0.25\linewidth}|p{0.25\linewidth}|p{0.25\linewidth}|p{0.25\linewidth}}
\textbf{QQP}     &\textbf{QNLI}        &\textbf{Wiki}    &\textbf{Books}           \\ \midrule
Nested quotes, Colors, Mathematics, Military conflict,
Population statistics, Relationship advice, School exam questions, Questions of comparison, Programming &
Biology, Geography, Technology, Numbers and dates, Military conflict, Population statistics, War history, Windows 8, Etymology &
Direct statement of fact, Music, Sporting, Age distribution, Television shows, Olympic facts, Legalese, Measurements, School districts &
Interpersonal relationships, Nature, Quoted speech, Spanish, Sentence fragments, Medieval Europe, Very long sentences, Flirtation
\end{tabular}}
\caption{\label{tab:annotations}Sample annotated patterns for each dataset.}
\end{table*}

\section{The Illusion}
\label{sec:illusion}

\begin{figure}[!b]
    \centering
    \includegraphics[width=0.8\columnwidth]{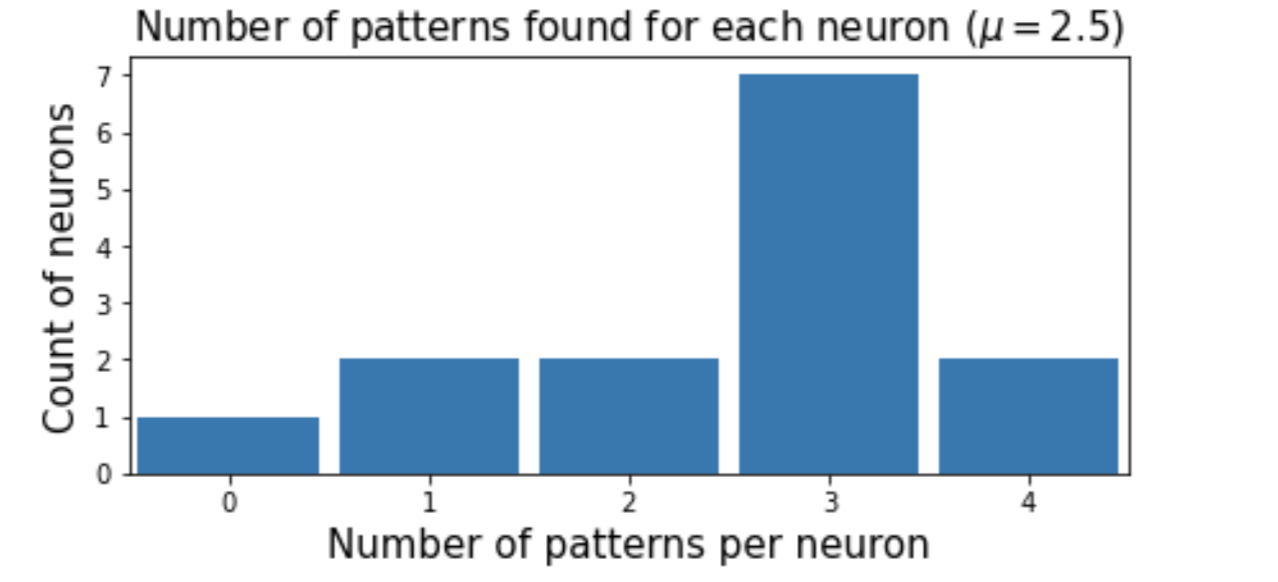} 
    \bigskip \\
    \begin{tabular}{l|ll}
    \toprule
    \textbf{Condition}  & \textbf{Mean}   & \textbf{Stdev}\\ 
    \midrule
    \textbf{Neurons} & 6.80 & 2.37 \\
    \textbf{Random Directions} & 6.89 & 2.13 \\ 
    \textbf{Random Sentences} & 5.05 & 1.96\\
    \bottomrule
    \end{tabular}
    \caption{Annotation statistics. (\textbf{Top}) The number of distinct patterns found for each annotated neuron across all datasets. We manually looked over the annotations and counted the number of unique patterns per neuron, grouping semantically equivalent annotations together. (\textbf{Bottom}) The number of sentences belonging to a pattern for the different experimental conditions for the sentence groups that are found to be meaningful. We required each pattern to have at least three positive examples. At most a pattern could have ten positive examples, because we only showed the top ten activating sentences for any given direction. \label{fig:distinct_patterns}}

\end{figure}

Each set of sentences was annotated by two annotators. Table \ref{tab:meaning_counts} shows how often both annotators found at least one pattern. \textit{Conflicting} indicates that one annotator found a pattern and the other did not. 
To establish a baseline, we also looked for patterns among randomly drawn sets of ten sentences. We found 14\% of random sets of sentences to contain patterns, suggesting that our datasets contain intrinsic topic biases. However, these biases are not sufficient to explain our results. We found that more than 80\% of top activating sentences contained patterns (Table \ref{tab:meaning_counts}). The patterns found among top activating sentences were also much stronger in that they contained more positive examples than those found in random sets of sentences (Figure \ref{fig:distinct_patterns}). Finally, annotators were more in agreement about whether top activating sentences were meaningful: only 8\% of annotators were split over whether a neuron was meaningful, and 18\% were split over whether a random direction was meaningful, compared to 38\% in the case of random sets of sentences.

Examples of our annotations are found in Table \ref{tab:annotations} (a full list can be found in the Appendix - Table \ref{tab:annotations_full}). Many of the patterns we found are quite general, for example that a set of sentences contain quoted speech, or that they concern nature. Nevertheless, the annotations of top activating sentences often changed dramatically depending on which dataset the sentences were drawn from. In fact, for each neuron we measured on average 2.5 \textit{distinct} patterns across QQP, QNLI, Wiki, and Books (Figure \ref{fig:distinct_patterns}).

Our results suggest that the illusion of meaningfulness we observed in neuron 221 (Section \ref{sec:introduction}) is a general, reproducible phenomenon of BERT. They also show that the illusion is equally pervasive among neurons and random directions. Based on an informal investigation of layers 2 and 7 we believe the illusion occurs for earlier layers as well, although a full analysis is beyond the scope of this paper. 

\section{Explaining the illusion}

What gives rise to this illusion? How could the same direction seemingly encode completely different concepts?

We propose that the illusion can be traced to three sources:
\begin{compactenum}
    \item Dataset idiosyncrasy
    \item Local semantic coherence in BERT's embedding space
    \item Annotator error
\end{compactenum}
Next, we discuss each source in detail.

\subsection{Datasets are idiosyncratic}   

\begin{figure}[!t]
    \centering
    \includegraphics[width=0.8\columnwidth]{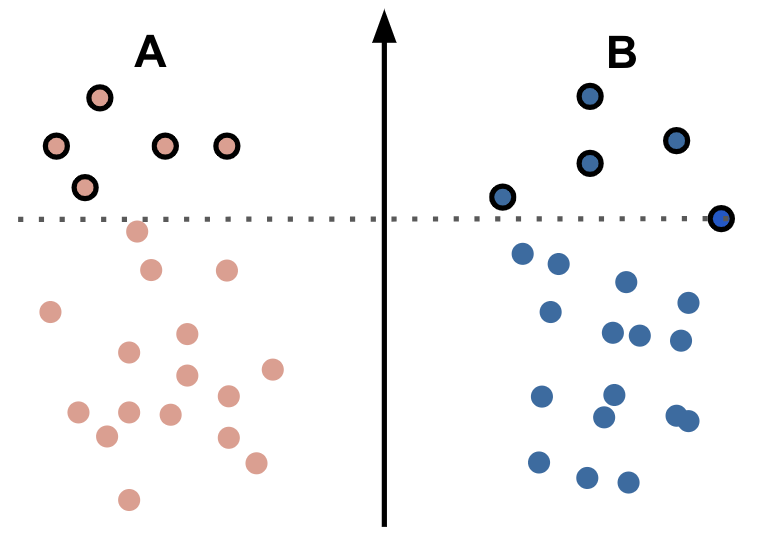}
    \caption{Schematic illustration of how top activating sentences from geometrically distinct datasets could be semantically unrelated. The arrow represents a direction in embedding space. Red dots represent sentences from dataset A, and blue dots represent sentences from dataset B. The dots outlined in black represent the the top activating sentences.\label{fig:dataset_schematic}}
\end{figure}

First, we consider the hypothesis:
\begin{hypothesis}
QQP, QNLI, Wiki, and Books occupy distinct regions of BERT's embedding space. 
\end{hypothesis}
This would contribute to the illusion because if the four datasets occupy non-overlapping slices of BERT's embedding space, then in any direction the top activating sentences from each dataset will come from distinct regions of the embedding space (Figure \ref{fig:dataset_schematic}). 

\textbf{Experiments} 

To test our hypothesis, we performed two experiments:
\begin{compactenum}
    \item We built an exploratory visualization of the QQP, QNLI, Wiki, and Books dataset embeddings using the UMAP dimensionality reduction algorithm. %
    \item We trained a linear SVM classifier to distinguish between the four datasets based on their sentence embeddings.
\end{compactenum}

\textbf{Results} 

\begin{figure}[!bht]
    \centering
    \includegraphics[width=0.6\columnwidth]{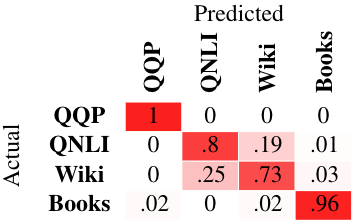}
    \caption{Confusion matrix for a linear classifier trained to separate QQP, QNLI, Wiki, and Books sentence embeddings. Most datasets are easily separable in the embedding space. \label{fig:confusion}}
\end{figure}

Our exploratory visualization shows that sentences cluster neatly by dataset (Figure \ref{fig:umap}). Our linear SVM is also able to distinguish between the datasets with high accuracy (Figure \ref{fig:confusion}). These results suggest that QQP, QNLI, Wiki and Books represent relatively idiosyncratic slices of language, thus the top activating sentences for a given neuron from one dataset do not necessarily resemble those from another dataset despite having similar activation values—looking at all the pairs of datasets across all neurons, 38\% of the ranges of the top 10 activations overlap. Our results align with previous work demonstrating that BERT representations can be used to disambiguate datasets \cite{goldberg2020}. However, while this explains why we would find different patterns in different datasets, why should there be patterns at all? We address this question in the next section.

\subsection{Local semantic coherence}
\label{sec:concept_def}

We hypothesize that another source of the illusion is local semantic coherence in BERT's embedding space geometry. 

Before testing this, we observe:

\begin{observation}
Top activating sentences manifest patterns from both local semantic coherence and global directions.
\end{observation}

This observation may seem obvious. However, suppose that a neuron encodes a certain concept: one might assume that its top activating sentences would clearly point to the encoded concept. Instead, we observe that the sentences may manifest patterns that do not match the encoded concept.

\begin{figure}[!tb]
    \centering
    \includegraphics[width=0.99\columnwidth]{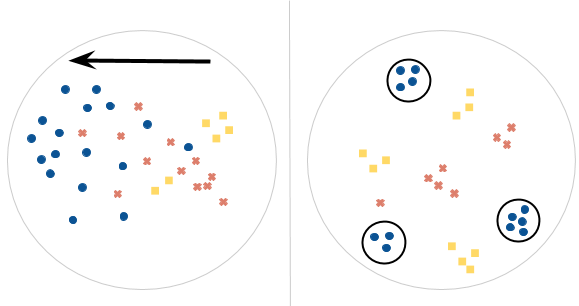}
    \caption{Global versus local concepts. \textbf{Left: Global concept illustration.} Blue circles represent sentences containing a global concept. As one moves in the concept direction indicated by the arrow, the density of sentences containing the global concept increases. \textbf{Right: Local concepts illustration.} The red, yellow, and blue shapes represent sentences containing three different concepts. Sentences containing the same concept cluster together, such that there is local concept coherence throughout the space. However, there is no global direction along which any particular concept becomes more dominant. \label{fig:global_concept_illustration}}
\end{figure}

Before further analysis, we describe three ways in which BERT may learn to represent concepts (Figure \ref{fig:global_concept_illustration}):

\textit{Global concepts:} Global concepts become increasingly prevalent as one moves through the embedding space along a linear trajectory. For example, if \textit{math} is a global concept, then there is a direction in the embedding space such that, starting from any point and moving along that direction, one will tend to find more and more sentences relating to math. Or if the concept is \textit{positivity}, words like \textit{happy, sunny, bliss, awesome} might increase in frequency. In our analysis we use the presence of certain tokens as a proxy for concepts. By extension, if a direction in embedding space is correlated with the occurrence of certain tokens, it suggests the existence of an underlying concept direction.

\textit{Dataset-level concepts:} Like global concepts, a \textit{dataset-level} concept is associated with a direction in the embedding space. However, unlike global concepts, this direction is only meaningful within the region of the embedding space where samples from the dataset tend to be found. Thus, dataset-level concepts do not generalize to arbitrary inputs. 

\textit{Local concepts:} On the other hand, \textit{local} concepts emerge only as clusters in the embedding space, and lack a direction. 
For example, suppose that \textit{math} is a local concept. Then if one looks in the neighborhood of the sentence $ e = m c^2 $, one may find many other sentences containing numbers and symbols because the model groups math-related sentences together. However, there will not be any particular direction along which such sentences become increasingly prevalent.

\begin{figure}[bh]
    \centering
    \includegraphics[width=0.99\columnwidth]{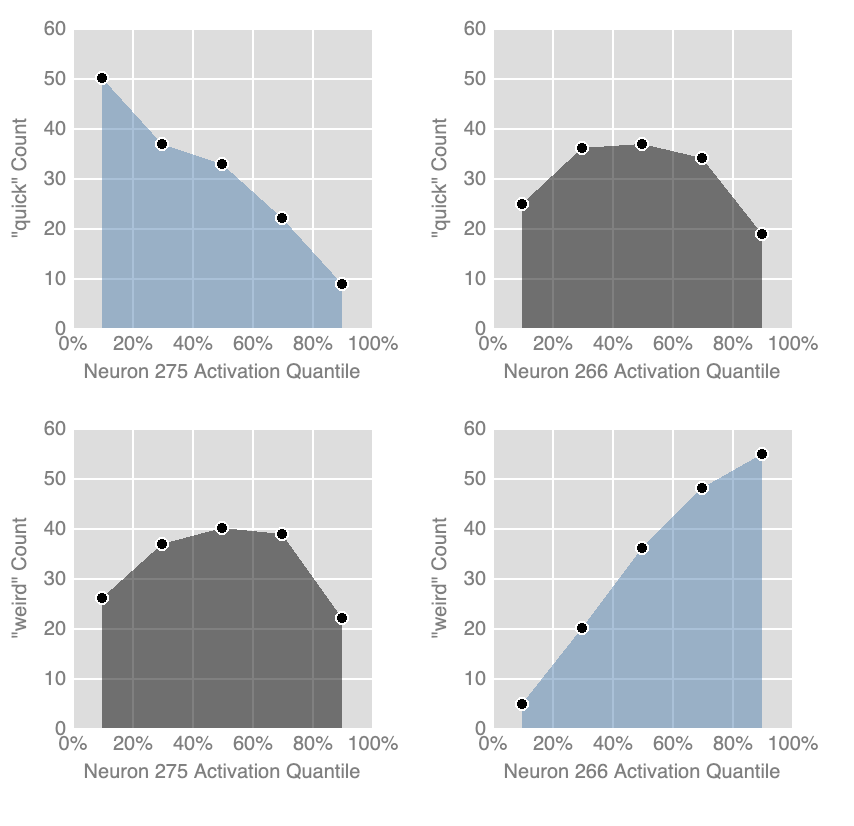}
    \caption{\textbf{(Top left)} The frequency of the word ``quick" monotonically decreases as we look at sentences that increasingly activate neuron 275 in QQP. This suggests that neuron 275 encodes a global concept (Section \ref{sec:concept_def}) that relates to ``quick". \textbf{(Top right)} Most neurons, such as neuron 266, do not correlate with the frequency of ``quick". These are illustrative examples, Table \ref{tab:monotonic-combinations} shows over a quarter of neuron/token pair are monotonic.
    \label{fig:global_concept}}
\end{figure}

\begin{table}[th]
\centering
\resizebox{\columnwidth}{!}{%
\renewcommand{\arraystretch}{1.4}
\begin{tabular}{@{}llllrrr@{}}
\multicolumn{4}{l}{\textbf{Datasets}} & \multicolumn{1}{l}{\textbf{Monotonic}} & \multicolumn{1}{l}{\textbf{Increasing}} & \multicolumn{1}{l}{\textbf{Decreasing}} \\ \midrule
Books &      &      &     & 27.0\% & 13.4\% & 13.6\% \\
      & QNLI &      &     & 22.7\% & 11.3\% & 11.4\% \\
      &      & Wiki &     & 29.6\% & 14.7\% & 15.0\% \\
      &      &      & QQP & 27.8\% & 13.9\% & 13.9\% \\
\rowcolor[HTML]{EFEFEF} 
Books & QNLI &      &     & 7.4\%  & 3.0\%  & 2.9\%  \\
\rowcolor[HTML]{EFEFEF} 
Books &      & Wiki &     & 9.9\%  & 4.3\%  & 4.3\%  \\
\rowcolor[HTML]{EFEFEF} 
      & QNLI & Wiki &     & 10.9\% & 5.3\%  & 5.4\%  \\
\rowcolor[HTML]{EFEFEF} 
Books &      &      & QQP & 9.2\%  & 3.9\%  & 3.9\%  \\
\rowcolor[HTML]{EFEFEF} 
      & QNLI &      & QQP & 7.7\%  & 3.2\%  & 3.2\%  \\
\rowcolor[HTML]{EFEFEF} 
      &      & Wiki & QQP & 9.8\%  & 4.0\%  & 4.1\%  \\
\rowcolor[HTML]{C0C0C0} 
Books & QNLI & Wiki &     & 4.2\%  & 1.8\%  & 1.8\%  \\
\rowcolor[HTML]{C0C0C0} 
Books & QNLI &      & QQP & 3.0\%  & 1.2\%  & 1.2\%  \\
\rowcolor[HTML]{C0C0C0} 
Books &      & Wiki & QQP & 3.9\%  & 1.7\%  & 1.6\%  \\
\rowcolor[HTML]{C0C0C0} 
      & QNLI & Wiki & QQP & 4.2\%  & 1.8\%  & 1.9\%  \\
\rowcolor[HTML]{9B9B9B} 
Books & QNLI & Wiki & QQP & 1.9\%  & 0.8\%  & 0.8\% 
\end{tabular}}
\caption{The table shows how often the token counts across quintiles of activations for each neuron/token pair are monotonically increasing or decreasing in each combination of datasets. Combinations with more datasets have darker backgrounds. There are some linear directions in the embedding space that are correlated with the same tokens across all datasets (for the 915 tokens appearing at least 100 times in each of the four datasets). \label{tab:monotonic-combinations}}
\end{table}

\begin{table}[ht]
\resizebox{\columnwidth}{!}{%
\begin{tabular}{p{0.99\columnwidth}}
\textbf{Tokens whose frequencies change monotonically:} \\
\hline
\textbf{"} (125), \textbf{can} (120), \textbf{is} (99), \textbf{are} (98), \textbf{was} (97), \textbf{that} (91), \textbf{if} (88), \textbf{were} (86), \textbf{to} (85), \textbf{would} (84), \textbf{a} (82), \textbf{it} (80), \textbf{they} (78), \textbf{not} (77), \textbf{god} (76), \textbf{for} (75), \textbf{which} (73), \textbf{more} (73), \textbf{she} (70), \textbf{of} (68)
\end{tabular}}
\caption{\label{tab:most_monotonic_tokens}This table lists the tokens from all four datasets that are most often monotonically changing across the embedding space. For each token, we indicate in parenthesis the number of neurons for which the token changes monotonically in frequency as one moves along the neuron axis.}
\end{table}

We find evidence for both global and dataset-level concept directions in BERT's embedding space. Figure \ref{fig:global_concept} illustrates the change in certain token frequencies as one moves along various neuron directions. Some tokens, such as ``weird'', monotonically change in frequency as one moves along neuron 266, while other token frequencies are unchanged. Table \ref{tab:monotonic-combinations} illustrates that this is not an isolated phenomenon, rather there are many tokens that monotonically increase or decrease along different neurons. Table \ref{tab:monotonic-combinations} further shows that these patterns can be unique to a particular dataset, suggesting the existence of dataset-level concept directions. We note that the baseline probability of a token (e.g. ``weird''ness) being monotonic with respect to a neuron is much lower than the measured rates\footnote{We use quintiles to measure monotonicity. The probability of a random set of quintiles being monotonic is 2/5! (1.7\%).}. Table \ref{tab:most_monotonic_tokens} shows the most monotonic tokens across datasets, hinting that BERT may have learned to encode certain pervasive concepts as global directions, e.g. pronouns and common verbs.

Despite the fact that global and dataset-level concept directions exist, we posit that they will be difficult to identify from top activating sentences. First, sentences typically engage with multiple concepts. Thus even if a neuron encodes a concept, its top activating sentences may have other concepts in common as well. %
Second, the directions we annotated are likely to themselves align with multiple concepts, and the top activating sentences for each concept may look unrelated when listed together. As an example, suppose we have a dataset in which every sentence represents exactly one concept and we pick a direction that is a linear combination of ten global concept directions. When we look at the ten sentences that most highly activate this direction, we may not see any patterns at all. In the extreme case, we may only see one sentence representing each concept, making it impossible to identify any patterns across sentences.

Formally we define a concept distribution as: $C = [c_1, c_2, ..., c_N]$ where $\sum_{i=1}^{N} c_i = 1.0$. The concept purity of a sentence or a direction is defined by the skew of its concept distribution vector. A sentence that aligns with exactly one concept would have a one-hot concept distribution vector.

\subsubsection{Measuring Local Concepts}
\label{sec:locality}

Having discussed the difficulty of identifying concept directions from top activating sentences, in this section we aim to show that:

\begin{hypothesis}
\label{hyp:measure_loc}
When annotating top activating sentences, people identify concepts emerging from local semantic coherence. 
\end{hypothesis}

This would explain the illusion because local concepts are not necessarily associated with a direction in the activation space, but rather with a semantic cluster. %

We propose the following analysis to test our hypothesis. The main intuition is that if a concept we annotate emerges from local semantic coherence, then for sentences manifesting the concept, their neighborhood in the original embedding space should look very similar to their neighborhood when projected onto the concept direction\footnote{Our analysis measures local coherence among top activating sentences. If both neighborhoods match, our analysis cannot determine whether the local coherence comes from a local or a global concept. However we know that for the directions whose meaning changes across datasets, the source of local coherence cannot be a global concept. On the other hand, a neighborhood mismatch would mean something besides local coherence is responsible for any pattern.}. 
Rather than measuring exact equality of sentences in the two neighborhoods (which is very sensitive), we compare their pairwise distance distributions.

More formally, let $N_k(s)$ be the $k$ nearest neighbors of a sentence $s$:
$$
N_k(s) = \argmax_{S^{\prime} \subset S, |S^{\prime}|=k} \: \sum_{\hat{s} \in S^{\prime}} e_{s} \cdot e_{\hat{s}}
$$
where $e_s$ denotes the embedding of a sentence in the original embedding space. Consistent with our annotation protocol we use $k=10$.

Let $S_{p,k}$ define the top activating $k$ sentences for a direction $p$. For each sentence $s$ in $S_{p,k}$, we measure the dot product between $s$ and its nearest neighbors $s^{\prime} \in N_{k}(s)$. We call this set of distances $D_{p, nearest}$. We then measure the dot product between all pairs of top activating sentences ($S_{p,k}$), calling this set $D_{p, top}$. Finally, we measure the dot product between each top activating sentence and $k$ random sentences in the dataset to establish a baseline, calling this set $D_{p, random}$. 

\begin{figure}[!htb]
  \centering
  \includegraphics[width=0.99\linewidth]{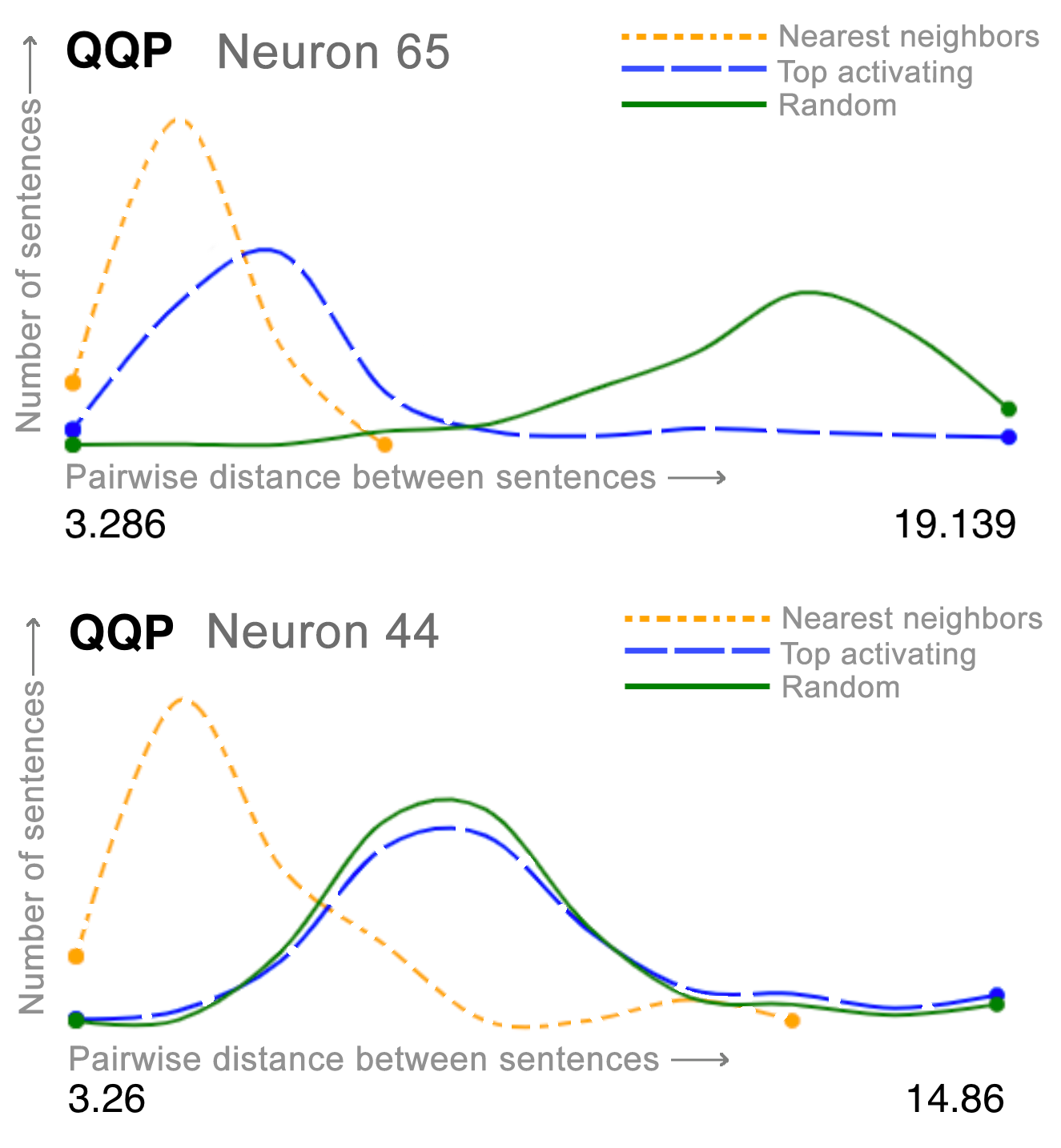}
  \caption{The distribution of distances in the original embedding space for QQP's top activating sentence for neurons 65, 44. Orange lines show the distances between the neurons' top activating sentence and its nearest neighbors in the original embedding space. Blue lines show the distances between the top activating sentence and other top activating sentences. Green lines show the distances between the top activating sentence and randomly drawn sentences. \textbf{(Top)} A neuron deemed meaningful by annotators. Observe that the blue line significantly overlaps with the orange line, suggesting that the meaning comes from the top activating sentence's local neighborhood. \textbf{(Bottom)} A neuron deemed meaningless by annotators. Observe the high degree of overlap between the distribution of distances between the top activating sentences and the distances between random sentences.\label{fig:locality_histograms}}
  \vspace{-0.1cm}
\end{figure}

The histograms in Figure \ref{fig:locality_histograms} shows how $D_{p, nearest}$, $D_{p, top}$, and $D_{p, random}$ compare. On top is a neuron our annotators deemed meaningful. The pairwise distances between the top activating sentences ($D_{p, top}$) overlap with those between the top activating sentences and their nearest neighbors in the original embedding space ($D_{p, nearest}$). By contrast, on the bottom is a meaningless neuron, where the pairwise distances between the top activating sentences ($D_{p, top}$) match those between the top activating and random sentences ($D_{p, random}$).

We extend this analysis across all the directions we annotated and propose a metric based on the Jaccard similarity:
$$
L(h_1, h_2) = \frac{\sum_i \min(h_1(i), h_2(i))}{\sum_i \max(h_1(i), h_2(i))}
$$
$L$ measures the intersection over union between any two histograms $h_1$ and $h_2$. We call $L$ the locality score for a direction when applied to the histograms of $D_{p, nearest}$ and $D_{p, top}$. If Hypothesis \ref{hyp:measure_loc} is true, $L$ should be high for meaningful directions and low for meaningless directions. 

Table \ref{tab:locality_correlation} shows that the locality scores for meaningless neurons are indeed significantly lower than for meaningful neurons. This means that the patterns found among top activating sentences arise primarily from \textit{local} geometry. 

\begin{table}[!htb]
\centering
\resizebox{\columnwidth}{!}{%
\begin{tabular}{llllll}
\toprule
    Locality (Mean)   & \textbf{All}   & \textbf{QQP}   & \textbf{QNLI}  & \textbf{Wiki}  & \textbf{Books}  \\ \midrule
Meaningful  & 0.026 & 0.012 & 0.014 & 0.037 & 0.042   \\
Meaningless & 0.010 & 0.003 & 0.008 & 0.016 & 0.012   \\
p-value     & 0.0004 & 0.009 & 0.090 & 0.064 & 0.014 \\ \bottomrule
\end{tabular}}
\caption{Per sentence locality scores (based on the histogram intersection over union measure).\label{tab:locality_correlation}}
\end{table}

\subsubsection{Analyzing outlier sentences}

We present our preceding analysis of local semantic coherence as general, but our experiment targets the edges of the activation space. Perhaps the properties of local semantic coherence are unique to a small set of outlier sentences which maximally activate across the directions we annotated.

We consider this possibility for the QQP dataset. Below are the three QQP sentences that maximally activate the most neurons:
\begin{compactitem}
    \item \textit{"what does snoop dogg mean by ""lolos"" in the line ""hit the corners in them lolos girl ""in dr. dre's 2001 hit ""still d.r.e""?"} (56 top activations)
    \item \textit{"what are some songs like ""the dying of the light - noel gallaghar""?"} (40 top activations)
    \item \textit{"what inspired the tv show ""skins""?"} (32 top activations)
\end{compactitem}

Taking the pairwise Euclidean distances between every QQP sentence, these three sentences have the highest mean distances. This pattern holds across the entire distribution: the 20 most distant QQP sentences all have at least 11 top activations and the most distant 1\% of sentences account for 48\% of all top activations.  

These numbers suggest a lack of diversity among QQP's outlier sentences. Indeed, looking at our experimental results, it appears that we annotated the same sentences across multiple directions: several neurons are maximally activated by nested quotation marks, several others by movie plots and several others by math equations. However most sentences are only top activating once or twice; across the 7,680 top ten activating sentences, there are 4,551 unique sentences. And the illusion \textit{persists} when the most distant 1\% and 10\% of sentences are omitted.

\subsection{Annotator error}
A final source of the illusion is our ability to see patterns where they may not exist. As noted in Section \ref{sec:illusion}, annotators often disagreed over whether a set of sentences constituted a pattern. Additionally, annotators differ significantly in their tendency to find patterns (Table \ref{tab:annotator_variance}). This suggests certain patterns may be less an objective property of the sentences than products of an individual annotator's (overactive) imagination.
\begin{table}[!htb]
\centering
\resizebox{0.99\columnwidth}{!}{%
\begin{tabular}{lll}
\toprule
\textbf{Annotator} & \textbf{Directions annotated} & \textbf{Patterns found}  \\ \midrule
0  & 30          & 27 (0.90) \\
1  & 33          & 27 (0.82) \\
2  & 24          & 23 (0.92) \\
3  & 40          & 23 (0.56) \\
4  & 52          & 30 (0.58) \\
5  & 25          & 40 (1.6) \\ \bottomrule
\end{tabular}}
\caption{\label{tab:annotator_variance} The total number of annotations and the number of patterns found per annotator. The number of patterns can exceed the number of annotations because a single direction may exhibit multiple patterns.}
\vspace{-0.6cm}
\end{table}

\section{Conclusion} 
In this work we investigated how BERT represents meaning by looking for patterns among top activating sentences along directions in the model's activation space.
In doing so, we uncovered a counterintuitive ``illusion'': seemingly consistent patterns that turn out to be contingent on the test dataset. We then identified several possible sources of this behavior: the idiosyncratic nature of widely used natural language datasets, the geometry of BERT's activation space, and annotator bias. Furthermore, despite these ``illusory'' findings, we also provide evidence that there do exist meaningful global directions in BERT's activation space.

Several avenues for future work suggest themselves. It would be interesting to replicate our analysis in other settings, such as with other sentence models, with token-level embeddings, and with earlier layers of BERT. It would also be natural to look for similar illusions in other types of input: images, graphs, and so forth. In any of these contexts, there is clearly much work to be done in characterizing the geometry of concept representation.

Our work also draws needed attention to the role of datasets in interpretability research. 
It is a classic problem in machine learning that a model trained on one dataset may not perform well on another. The results we have described illustrate something slightly different: how an \textit{interpretation} that seems valid for one dataset may not generalize to other contexts. Examining how interpretations function across different data distributions may be a helpful step in establishing their validity.

\textbf{Acknowledgements}
We would like to thank Ian Tenney, Jasmijn Bastings, Katherine Lee and Been Kim for their reviews and helpful discussions during the process of writing this paper.

\bibliography{illusion}

\begin{thebibliography}{29}
\providecommand{\natexlab}[1]{#1}
\providecommand{\url}[1]{\texttt{#1}}
\expandafter\ifx\csname urlstyle\endcsname\relax
  \providecommand{\doi}[1]{doi: #1}\else
  \providecommand{\doi}{doi: \begingroup \urlstyle{rm}\Url}\fi

\bibitem[Aharoni \& Goldberg(2020)Aharoni and Goldberg]{goldberg2020}
Aharoni, R. and Goldberg, Y.
\newblock Unsupervised domain clusters in pretrained language models.
\newblock In \emph{Association for Computational Linguistics}, pp.\
  7747--7763, 2020.

\bibitem[Bau et~al.(2017)Bau, Zhou, Khosla, Oliva, and
  Torralba]{bau2017network}
Bau, D., Zhou, B., Khosla, A., Oliva, A., and Torralba, A.
\newblock Network dissection: Quantifying interpretability of deep visual
  representations.
\newblock In \emph{Proceedings of the IEEE conference on computer vision and
  pattern recognition}, pp.\  6541--6549, 2017.

\bibitem[Bengio et~al.(2003)Bengio, Ducharme, Vincent, and
  Janvin]{bengio2003neural}
Bengio, Y., Ducharme, R., Vincent, P., and Janvin, C.
\newblock A neural probabilistic language model.
\newblock \emph{The journal of machine learning research}, 3:\penalty0
  1137--1155, 2003.

\bibitem[Bolukbasi \& Chang(2016)Bolukbasi and Chang]{bolukbasi2016}
Bolukbasi, T. and Chang, K.-W.
\newblock Man is to computer programmer as woman is to homemaker? debiasing
  word embeddings.
\newblock In \emph{Proceedings of the 30th International Conference on Neural
  Information Processing Systems}, pp.\  4356--4364, 2016.

\bibitem[Bäuerle \& Wexler()Bäuerle and Wexler]{bauerle_wexler}
Bäuerle, A. and Wexler, J.
\newblock What does bert dream of?
\newblock URL
  \url{https://pair-code.github.io/interpretability/text-dream/explainable/}.

\bibitem[Dalvi et~al.(2019)Dalvi, Durrani, and Sajjad]{dalvi2019}
Dalvi, F., Durrani, N., and Sajjad, H.
\newblock What is one grain of sand in the desert? analyzing individual neurons
  in deep nlp models.
\newblock In \emph{Proceedings of the 33rd AAAI Conference on Artificial
  Intelligence}, pp.\  6309--6317, 2019.

\bibitem[Devlin et~al.(2018)Devlin, Chang, Lee, and Toutanova]{devlin2018bert}
Devlin, J., Chang, M.-W., Lee, K., and Toutanova, K.
\newblock Bert: Pre-training of deep bidirectional transformers for language
  understanding.
\newblock \emph{arXiv preprint arXiv:1810.04805}, 2018.

\bibitem[Durrani et~al.(2020)Durrani, Dalvi, Sajjad, and Belinkov]{dalvi2020b}
Durrani, N., Dalvi, F., Sajjad, H., and Belinkov, Y.
\newblock Analyzing individual neurons in pre-trained language models.
\newblock 2020.
\newblock URL \url{https://arxiv.org/pdf/2010.02695.pdf}.

\bibitem[Iyer et~al.(2017)Iyer, Dandekar, and Csernai]{WinNT}
Iyer, S., Dandekar, N., and Csernai, K.
\newblock First quora dataset release: Question pairs, 2017.
\newblock URL
  \url{https://data.quora.com/First-Quora-Dataset-Release-Question-Pairs}.

\bibitem[Kaneko \& Bollegala(2019)Kaneko and
  Bollegala]{DBLP:journals/corr/abs-1906-00742}
Kaneko, M. and Bollegala, D.
\newblock Gender-preserving debiasing for pre-trained word embeddings.
\newblock \emph{CoRR}, abs/1906.00742, 2019.
\newblock URL \url{http://arxiv.org/abs/1906.00742}.

\bibitem[Kim et~al.(2018)Kim, Wattenberg, Gilmer, Cai, Wexler, Viegas,
  et~al.]{kim2018interpretability}
Kim, B., Wattenberg, M., Gilmer, J., Cai, C., Wexler, J., Viegas, F., et~al.
\newblock Interpretability beyond feature attribution: Quantitative testing
  with concept activation vectors (tcav).
\newblock In \emph{International conference on machine learning}, pp.\
  2668--2677, 2018.

\bibitem[Li et~al.(2015)Li, Chen, Hovy, and
  Jurafsky]{DBLP:journals/corr/LiCHJ15}
Li, J., Chen, X., Hovy, E.~H., and Jurafsky, D.
\newblock Visualizing and understanding neural models in {NLP}.
\newblock \emph{CoRR}, abs/1506.01066, 2015.
\newblock URL \url{http://arxiv.org/abs/1506.01066}.

\bibitem[Manning et~al.(2020)Manning, Hewitt, Clark, Khandelwal, and
  Levy]{manning2020}
Manning, C., Hewitt, J., Clark, K., Khandelwal, U., and Levy, O.
\newblock Emergent linguistic structure in artificial neural networks trained
  by self-supervision.
\newblock \emph{Proceedings of the National Academy of Sciences of the United
  States of America}, 2020.

\bibitem[Manzini et~al.(2019)Manzini, Lim, Tsvetkov, and
  Black]{DBLP:journals/corr/abs-1904-04047}
Manzini, T., Lim, Y.~C., Tsvetkov, Y., and Black, A.~W.
\newblock Black is to criminal as caucasian is to police: Detecting and
  removing multiclass bias in word embeddings.
\newblock \emph{CoRR}, abs/1904.04047, 2019.
\newblock URL \url{http://arxiv.org/abs/1904.04047}.

\bibitem[Mikolov et~al.(2013)Mikolov, Chen, Corrado, and
  Dean]{mikolov2013efficient}
Mikolov, T., Chen, K., Corrado, G., and Dean, J.
\newblock Efficient estimation of word representations in vector space, 2013.

\bibitem[Na et~al.(2019)Na, Choe, Lee, and Kim]{na2019}
Na, S., Choe, Y.~J., Lee, D.-H., and Kim, G.
\newblock Discovery of natural language concepts in individual units of cnns.
\newblock 2019.
\newblock In the Proceedings of ICLR.

\bibitem[Nguyen et~al.(2016)Nguyen, Dosovitskiy, Yosinski, Brox, and
  Clune]{nguyen2016}
Nguyen, A., Dosovitskiy, A., Yosinski, J., Brox, T., and Clune, J.
\newblock Synthesizing the preferred inputs for neurons in neural networks via
  deep generator networks.
\newblock In \emph{Advances in Neural Information Processing Systems}, pp.\
  3387--3395, 2016.

\bibitem[Olah et~al.(2017)Olah, Mordvintsev, and Schubert]{olah2017feature}
Olah, C., Mordvintsev, A., and Schubert, L.
\newblock Feature visualization.
\newblock \emph{Distill}, 2017.
\newblock \doi{10.23915/distill.00007}.
\newblock https://distill.pub/2017/feature-visualization.

\bibitem[Olah et~al.(2020)Olah, Cammarata, Schubert, Goh, Petrov, and
  Carter]{olah2020zoom}
Olah, C., Cammarata, N., Schubert, L., Goh, G., Petrov, M., and Carter, S.
\newblock Zoom in: An introduction to circuits.
\newblock \emph{Distill}, 2020.
\newblock \doi{10.23915/distill.00024.001}.
\newblock https://distill.pub/2020/circuits/zoom-in.

\bibitem[Poerner et~al.(2018)Poerner, Roth, and
  Sch{\"u}tze]{poerner-etal-2018-interpretable}
Poerner, N., Roth, B., and Sch{\"u}tze, H.
\newblock Interpretable textual neuron representations for {NLP}.
\newblock In \emph{Proceedings of the 2018 {EMNLP} Workshop {B}lackbox{NLP}:
  Analyzing and Interpreting Neural Networks for {NLP}}, pp.\  325--327,
  Brussels, Belgium, November 2018. Association for Computational Linguistics.
\newblock \doi{10.18653/v1/W18-5437}.
\newblock URL \url{https://www.aclweb.org/anthology/W18-5437}.

\bibitem[Raghu et~al.(2017)Raghu, Gilmer, Yosinski, and
  Sohl-Dickstein]{raghu2017svcca}
Raghu, M., Gilmer, J., Yosinski, J., and Sohl-Dickstein, J.
\newblock Svcca: Singular vector canonical correlation analysis for deep
  learning dynamics and interpretability, 2017.

\bibitem[Reimers \& Gurevych(2019)Reimers and Gurevych]{reimers2019}
Reimers, N. and Gurevych, I.
\newblock Sentence-bert: Sentence embeddings using siamese bert-networks.
\newblock In \emph{Proceedings of the 2019 Conference on Empirical Methods in
  Natural Language Processing and the 9th International Joint Conference on
  Natural Language Processing}, pp.\  3982--–3992, 2019.

\bibitem[Szegedy et~al.(2014)Szegedy, Zaremba, Sutskever, Bruna, Erhan,
  Goodfellow, and Fergus]{szegedy2014}
Szegedy, C., Zaremba, W., Sutskever, I., Bruna, J., Erhan, D., Goodfellow, I.,
  and Fergus, R.
\newblock Intriguing properties of neural networks.
\newblock In \emph{International Conference on Learning Representations}, 2014.
\newblock URL \url{http://arxiv.org/abs/1312.6199}.

\bibitem[Tenney et~al.(2019)Tenney, Xia, Chen, Wang, Poliak, McCoy, Kim, Durme,
  Bowman, Das, and Pavlick]{tenney2019what}
Tenney, I., Xia, P., Chen, B., Wang, A., Poliak, A., McCoy, R.~T., Kim, N.,
  Durme, B.~V., Bowman, S., Das, D., and Pavlick, E.
\newblock What do you learn from context? probing for sentence structure in
  contextualized word representations.
\newblock In \emph{International Conference on Learning Representations}, 2019.

\bibitem[Vig et~al.(2020)Vig, Gehrmann, Belinkov, Qian, Nevo, Sakenis, Huang,
  Singer, and Shieber]{vig2020causal}
Vig, J., Gehrmann, S., Belinkov, Y., Qian, S., Nevo, D., Sakenis, S., Huang,
  J., Singer, Y., and Shieber, S.
\newblock Causal mediation analysis for interpreting neural nlp: The case of
  gender bias, 2020.

\bibitem[Wang et~al.(2019)Wang, Singh, Michael, Hill, Levy, and
  Bowman]{wang2019glue}
Wang, A., Singh, A., Michael, J., Hill, F., Levy, O., and Bowman, S.~R.
\newblock {GLUE}: A multi-task benchmark and analysis platform for natural
  language understanding.
\newblock 2019.
\newblock In the Proceedings of ICLR.

\bibitem[Zeiler \& Fergus(2014)Zeiler and Fergus]{zeiler2014visualizing}
Zeiler, M.~D. and Fergus, R.
\newblock Visualizing and understanding convolutional networks.
\newblock In \emph{European conference on computer vision}, pp.\  818--833.
  Springer, 2014.

\bibitem[Zhou et~al.(2015)Zhou, Khosla, Lapedriza, Oliva, and
  Torralba]{zhou2015}
Zhou, B., Khosla, A., Lapedriza, A., Oliva, A., and Torralba, A.
\newblock Object detectors emerge in deep scene cnn.
\newblock In \emph{International Conference on Learning Representations}, 2015.

\bibitem[Zhu et~al.(2015)Zhu, Kiros, Zemel, Salakhutdinov, Urtasun, Torralba,
  and Fidler]{zhu2015aligning}
Zhu, Y., Kiros, R., Zemel, R., Salakhutdinov, R., Urtasun, R., Torralba, A.,
  and Fidler, S.
\newblock Aligning books and movies: Towards story-like visual explanations by
  watching movies and reading books, 2015.

\end{thebibliography}
\bibliographystyle{icml2020}

\clearpage
\appendix

\begin{table*}[!hbt]
\centering
\resizebox{\linewidth}{!}{%
\begin{tabular}{p{0.25\linewidth}|p{0.25\linewidth}|p{0.25\linewidth}|p{0.25\linewidth}}
\textbf{QQP}     &\textbf{QNLI}        &\textbf{Wiki}    &\textbf{Books}           \\ \midrule
School exam questions, High school math and physics, Comparison questions, Indian jobs, Jobs for students, Large body physics, Movies, Music, Math, Physics, Indian cities, Race, Science, Technology and consulting, Containing multiple quoted phrases, Hygiene, Tech support, Drugs, India, Programming, Mathemtical and chemical formulas, Cosmology, Containing quoted subjects, Internet companies, Corporate jargon, Quoted songs, Indian paperwork &
Military conflict, Ethnic groups (language, culture, history), Numbers and dates, Standards, Named places (state, university, etc.), DNA, Text book material, Population statistics / census, War history, Military moves, Geographic locations, Africa (mostly Somalia), Engineering, Governmental power, Geography, Language / linguistics, Group theory, Politics, Military, Governments and leaders and revolutions, Displaced people and slavery, Animals, Science, Renovations and infrastructure changes, US southern cities and especially in NC, Countries, Bible quotes and locations, Resettling and boundaries, War, Histories of ethnic groups, Math, Census data, Military maneuvers, Somalia / Eritrea, Law and legislative bodies, Historical events, Etymology, Windows 8, Mathematical groups, Years, Political conflict and revolution, Political history, Animals, Family, Geography, Ecology, Census, Chemistry, Weather history, Municipal facts, Indian, Demographics, Weather, Raleigh &
Settlement history, Direct statements of fact, Named locations, Voting stats of cities, Birth family occupations of individuals, Locations, British history, Age distribution, Census data, TV shows, Olympians, Daughters, War, Law / contracts, International sports, Media, School districts, National borders, Music, Properties of villages, Anatomical descriptions, Indiginous communities, Chemistry, Printing presses, Math, Statistics, Television shows, Legalese, French / German municipalities, Measurements, Rules, Minerals, Soccer, Medicine, School rules, Bands &
Spanish, Description then talking, Snippets describing people, Fantasy travel, Quotes, Body parts, Plans, Person doing a small physical action (laughing, rubbing hands, etc), French or Spanish, Different forms of "what\'s up? how\'s it going?", Sentence fragments (long noun phrases), Short statements about a character performing an action, About medieval European islands / cold places, Long blocks of quoted text, Short fragments separated by commas, Long blocks of text, Flirtation, Military, Non-English, Questions, Violence
\end{tabular}}
\caption{\label{tab:annotations_full}Annotated patterns for each dataset. Edited for brevity, duplicates removed.}
\end{table*}
\clearpage
\section{Normalization} 
We decided to conduct our analysis on raw embeddings, which are un-normalized. We found that the top activating sentences for directions in the embedding space were in general the same with and without normalization, therefore we decided to keep the vector space as close to how the neural network utilizes it as possible by not normalizing.

In addition, we observed that most vector norms are concentrated around $\sim14$ which is about $sqrt(768)/2$. This is about the norm of a vector which has all elements equal to 0.5. We also observed that the outputs of the neurons typically fall between -1.0 and 1.0.

\end{document}